\documentclass[]{interact}

\usepackage{epstopdf}
\usepackage{subfigure}
\usepackage{tabularx}
\usepackage{graphicx}
\usepackage{amsmath}
\usepackage{amssymb}
\usepackage{booktabs}
\usepackage{placeins} 
\usepackage{float} 
\usepackage{enumitem}
\usepackage{tabularx} 
\usepackage{longtable} 
\usepackage{array} 
\usepackage{ragged2e} 
\usepackage{booktabs} 
\usepackage{placeins}
\usepackage{afterpage} 
\usepackage{caption} 
\usepackage{booktabs} 
\usepackage{adjustbox} 

\begin{document}

\title{Spatial-Geometry Enhanced 3D Dynamic Snake Convolutional Neural Network for Hyperspectral Image Classification}

\author{
\name{Guandong Li\textsuperscript{a}\thanks{CONTACT Guandong Li. Email: leeguandon@gmail.com} and Mengxia Ye\textsuperscript{b}}
\affil{\textsuperscript{a}iFLYTEK, Shushan, Hefei, Anhui, China; \textsuperscript{b}Aegon THTF,Qinghuai,Nanjing,Jiangsu,China}
}

\maketitle

\begin{abstract}
Deep neural networks face several challenges in hyperspectral image classification, including complex and sparse ground object distributions, small clustered structures, and elongated multi-branch features that often lead to missing detections. To better adapt to ground object distributions and achieve adaptive dynamic feature responses while skipping redundant information, this paper proposes a Spatial-Geometry Enhanced 3D Dynamic Snake Network (SG-DSCNet) based on an improved 3D-DenseNet model. The network employs Dynamic Snake Convolution (DSCConv), which introduces deformable offsets to enhance kernel flexibility through constrained self-learning, thereby improving regional perception of ground objects. Additionally, we propose a multi-view feature fusion strategy that generates multiple morphological kernel templates from DSCConv to observe target structures from different perspectives and achieve efficient feature fusion through summarizing key characteristics. This dynamic approach enables the model to focus more flexibly on critical spatial structures when processing different regions, rather than relying on fixed receptive fields of single static kernels. The DSC module enhances model representation capability through dynamic kernel aggregation without increasing network depth or width. Experimental results demonstrate superior performance on the IN, UP, and KSC datasets, outperforming mainstream hyperspectral classification methods.
\end{abstract}


\section{Introduction}

Hyperspectral remote sensing images (HSI) play a crucial role in spatial information applications due to their unique narrow-band imaging characteristics. These images typically contain dozens to hundreds of continuous spectral bands, enabling precise capture of target spectral characteristics and significantly improving ground object recognition accuracy \cite{chang2003hyperspectral}. The imaging equipment synchronously records both spectral and spatial position information, integrating them into a three-dimensional data structure containing two-dimensional space and one-dimensional spectrum. As an important application of remote sensing technology, ground object classification demonstrates broad value in ecological assessment, transportation planning, agricultural monitoring, land management, and geological surveys \cite{bing2011intelligent}.

HSI faces a significant challenge called "spatial variability," where the same object exhibits different characteristics across regions due to atmospheric interference, lighting angles, and other imaging factors. Mixed categories may lead to inter-class similarity, where different classes share similar spectral features, increasing classification difficulty. Deep learning methods for HSI classification \cite{sun2016random, wang2019domain, gong2019cnn, safari2020multiscale,li2019doubleconvpool,li2018scene} have made substantial progress. In \cite{lee2017going} and \cite{zhao2016spectral}, Principal Component Analysis (PCA) was first applied to reduce dimensionality before extracting spatial information using 2D CNN. Methods like 2D-CNN \cite{makantasis2015deep, chen2016deep} require separate extraction of spatial and spectral features, failing to fully utilize joint spatial-spectral information while needing complex preprocessing. \cite{wang2018fast} proposed a Fast Dense Spectral-Spatial Convolutional Network (FDSSC) using serially connected 1D-CNN and 3D-CNN dense blocks. FSKNet \cite{li2022faster} introduced a 3D-to-2D module and selective kernel mechanism, while 3D-SE-DenseNet \cite{li2020hyperspectral} incorporated the SE mechanism into 3D-CNN to correlate feature maps across channels, activating effective information while suppressing ineffective features. DGCNet \cite{li2023dgcnet} designed dynamic grouped convolution (DGC) on 3D kernels, where each group includes small feature selectors to dynamically determine input channel connections based on activations, enabling CNNs to learn rich feature representations. DACNet \cite{li2025efficient} introduced dynamic attention convolution using multiple parallel kernels with attention weights, achieving adaptive spatial-spectral responses. DHSNet \cite{liu2025dual} proposed a Center Feature Attention-Aware Convolution (CFAAC) module that focuses on cross-scene invariant features, enhancing generalization capability. LGCNet \cite{li2025spatial} improved grouped convolution by designing learnable 3D convolution groups where both input channels and kernel groups can be learned, allowing flexible structures with better representation. Therefore, 3D-CNN has become a primary approach for HSI classification.

To leverage both CNN and Transformer advantages, many studies combine them to utilize local and global features. \cite{sun2022spectral} proposed a Spectral-Spatial Feature Tokenization Transformer (SSFTT) using 3D/2D convolutional layers for shallow features and Gaussian-weighted tokens in transformer encoders for high-level semantics. Some Transformer-based methods \cite{hong2021spectralformer} employ grouped spectral embedding and transformer encoders, but these treat spectral bands or spatial patches as tokens, causing significant redundant computations. Given HSI's inherent redundancy, these methods often underperform 3D-CNN approaches while requiring greater complexity.

3D-CNN can simultaneously sample spatial and spectral dimensions while preserving 2D convolution's spatial feature extraction capability. Although directly processing high-dimensional data eliminates preliminary dimensionality reduction, introducing spectral dimensions dramatically increases kernel parameters. Current methods like DFAN \cite{zhang2020deep}, MSDN \cite{zhang2019multi}, 3D-DenseNet \cite{zhang2019three}, and 3D-SE-DenseNet use dense connections that directly link each layer to all preceding layers, enabling feature reuse but introducing redundancy when later layers don't need early features. Therefore, efficiently enhancing 3D kernel representation while dynamically filtering redundant information in dense connections has become crucial for HSI classification.

HSI exhibits two distinct characteristics: (1) Many clustered regional features where some objects occupy minimal image proportions, making them vulnerable to complex background interference and prediction omissions; (2) Complex sparse structures showing separation and multiple branching morphologies that may exceed network receptive fields, causing recognition errors. Addressing these characteristics, we propose a Spatial-Geometry Enhanced 3D Snake Network (SG-DSCNet) that adaptively focuses on complex structures and elongated curved local features through DSCConv's deformability to capture long-range dependencies and non-local similarity. Unlike deformable convolution that freely learns geometric changes (potentially causing perception field wandering), our DSCConv considers multi-branch elongation and clustered region characteristics by introducing constrained offset learning. From 3D convolution's joint spatial-spectral perspective, we also propose a multi-view fusion strategy generating multiple morphological kernels to observe targets from different perspectives and fuse key features efficiently. This dynamic approach flexibly focuses on critical spatial structures without relying on fixed receptive fields.

Compared to traditional 3D-CNN, our method enhances spatial feature representation through geometry-enhanced dynamic kernels and fusion mechanisms without increasing network depth/width, particularly effective for uneven distributions or complex patterns. In spectral dimensions, DSCConv reduces redundant spectral aggregation, avoiding uniform sampling treatment in traditional 3D-CNN and alleviating parameter redundancy from high spectral dimensionality.

The main contributions are:

1. We propose a spatial-enhanced dynamic snake convolution network improving 3D-DenseNet for joint spatial-spectral HSI classification, addressing small clustered regions and elongated multi-branch scenarios through DSCConv and multi-view fusion, achieving strong performance on IN and UP datasets.

2. We introduce dynamic kernel mechanisms in 3D-CNN by modifying standard 3DConv with deformable offsets, implementing dynamic iteration and multi-view fusion to generate morphological kernel templates that effectively skip 3D-CNN redundancy.

3. SG-DSCNet is more concise than complex DL combination networks, requiring less computation while enhancing representation through spatial-enhanced dynamic kernel aggregation.

\section{Spatial-Geometry Enhanced 3D Dynamic Snake CNN}

\subsection{Dynamic Convolution}

Given HSI's sparse and clustered features, most methods focus on enhancing spatial-spectral feature extraction, where dynamic convolution offers an effective solution by strengthening kernel-level representation. LGCNet designed learnable group structures where input channels and kernel groups can be learned end-to-end. DGCNet introduced dynamic grouped convolution with small feature selectors per group. DACNet employed dynamic attention convolution using SE-generated weights with multiple parallel kernels. These approaches share dynamic convolution principles to address HSI's spatial-spectral complexity. Targeting 3D-CNN's redundancy, we propose DSCConv with constrained self-learning that fundamentally solves missing detection issues.

\subsection{3D Snake Convolution and Multi-View Fusion}

Hyperspectral imagery suffers from limited sample availability and exhibits sparse ground object characteristics with uneven spatial distribution, while containing substantial high-dimensional redundant information in the spectral domain. Although 3D-CNN architectures can leverage joint spatial-spectral information, how to achieve more effective deep extraction of spatial-spectral features remains a critical research challenge.As the core component of convolutional neural networks, convolutional kernels are typically regarded as information aggregators that combine spatial information and feature-dimensional data within local receptive fields. Composed of multiple convolutional layers, nonlinear layers, and downsampling layers, CNNs can capture image features from global receptive fields for comprehensive image representation. However, training high-performance networks remains challenging, with numerous studies focusing on spatial dimension improvements. For instance, residual structures enhance deep feature extraction by fusing outputs from different blocks, while DenseNet improves feature reuse through dense connections.

In feature extraction, 3D-CNNs process both spatial and spectral information of hyperspectral imagery through convolutional operations. However, their kernels often contain numerous redundant weights that contribute minimally to the final output. This redundancy is particularly prominent in joint spatial-spectral feature extraction:From the spatial perspective, the sparse and uneven distribution of ground objects may lead convolutional kernels to capture numerous irrelevant or low-information regions within local receptive fields. Spectrally, hyperspectral data typically contains hundreds of bands with high inter-band correlation and redundancy, making it difficult for convolutional kernels to effectively focus on discriminative features along the spectral axis. The spectral redundancy causes many parameters to merely serve as "fillers" in high-dimensional processing, failing to fully exploit deep patterns in joint spatial-spectral information. This weight redundancy not only increases computational complexity but may also impair the model's capability to represent sparse objects and complex spectral signatures, ultimately limiting 3D-CNN performance in hyperspectral image analysis.
This paper introduces geometrically enhanced deformable separable convolution (DSC Conv) and multi-view fusion for hyperspectral image processing. DSC Conv enhances kernel flexibility through learnable offset parameters, where self-adaptive offsets and constrained iterative optimization enable better focus on small clusters and elongated ground objects. The multi-view fusion strategy employs multiple morphological kernel templates to observe target structures from diverse perspectives, achieving comprehensive feature fusion.Spatially, our approach emphasizes responses to critical structural patterns, while spectrally implementing redundant band identification and adaptive skipping, thereby achieving effective spatial-spectral feature representation. This proves particularly advantageous for sparse ground objects in hyperspectral data, where limited samples (small-sample scenario) lead to significant feature map variations across different kernels. By integrating DSC Conv with 3D-DenseNet's hierarchical characteristics, we enable more effective feature extraction.

\begin{figure}[h]
\centering
\includegraphics[width=0.9\linewidth]{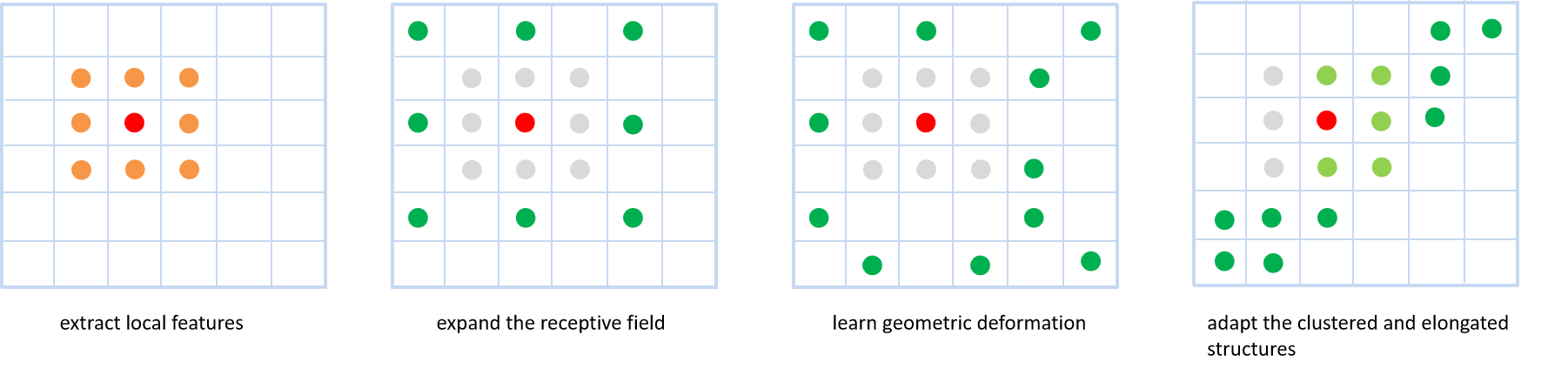}
\caption{Comparison between standard convolution, dilated convolution, deformable convolution, and DSC Conv}
\label{fig:conv_comparison}
\end{figure}

\subsubsection{Dynamic Snake Convolution}
In this section, we discuss how to implement Dynamic Snake Convolution (DSConv) for extracting local features of tubular structures. Given a standard 2D convolution coordinate set $K$ with center coordinate $K_i = (x_i, y_i)$. A $3 \times 3$ convolution kernel $K$ is represented as:

\begin{equation}
K = \{(x - 1, y - 1), (x - 1, y), \cdots, (x + 1, y + 1)\}
\end{equation}

To enhance the kernel's flexibility for capturing complex geometric features of targets, we introduce deformation offsets $\Delta$. However, if the model learns deformation offsets completely freely, the receptive field often deviates from the target, especially when processing small clusters and multi-branched elongated tubular structures. Therefore, we adopt an iterative strategy that sequentially selects the next position to process, ensuring attention continuity and preventing excessive diffusion of the receptive field due to large deformation offsets.

In Dynamic Snake Convolution, we linearize the standard convolution kernel along both x-axis and y-axis directions. Considering a kernel of size 9, taking the x-axis direction as an example, each grid position in $K$ is expressed as $K_{i\pm c} = (x_{i\pm c}, y_{i\pm c})$, where $c = 0, 1, 2, 3, 4$ represents the horizontal distance from the center grid. The selection of each grid position $K_{i\pm c}$ in kernel $K$ is a cumulative process. Starting from the center position $K_i$, positions farther from the center depend on their predecessor's position: $K_{i+1}$ adds an offset $\Delta = \{\delta|\delta \in [-1, 1]\}$ relative to $K_i$. Thus, the offsets require accumulation $\Sigma$ to ensure the kernel conforms to linear morphological structures. The x-axis transformation is:

\begin{equation}
K_{i\pm c} = 
\begin{cases}
(x_{i+c}, y_{i+c}) = (x_i + c, y_i + \Sigma_i^{i+c} \Delta_y), \\
(x_{i-c}, y_{i-c}) = (x_i - c, y_i + \Sigma_{i-c}^i \Delta_y),
\end{cases}
\end{equation}

The y-axis transformation is:

\begin{equation}
K_{j\pm c} = 
\begin{cases}
(x_{j+c}, y_{j+c}) = (x_j + \Sigma_j^{j+c} \Delta_x, y_j + c), \\
(x_{j-c}, y_{j-c}) = (x_j + \Sigma_{j-c}^j \Delta_x, y_j - c),
\end{cases}
\end{equation}

Since offsets $\Delta$ are typically fractional while coordinates are integers, we employ bilinear interpolation:

\begin{equation}
K = \sum_{K'} B(K', K) \cdot K'
\end{equation}

where $K$ represents fractional positions from Equations (2) and (3), $K'$ enumerates all integer spatial positions, and $B$ is the bilinear interpolation kernel, which can be decomposed into two one-dimensional kernels:

\begin{equation}
B(K, K') = b(K_x, K'_x) \cdot b(K_y, K'_y)
\end{equation}

Due to two-dimensional (x-axis, y-axis) variations, our Dynamic Snake Convolution kernel covers a $9 \times 9$ selectable receptive field during deformation. The DSConv kernel aims to better adapt to small clusters and elongated multi-branch spreading structures based on dynamic structures, enabling more effective perception of key features.

\begin{figure}[h]
\centering
\includegraphics[width=0.9\linewidth]{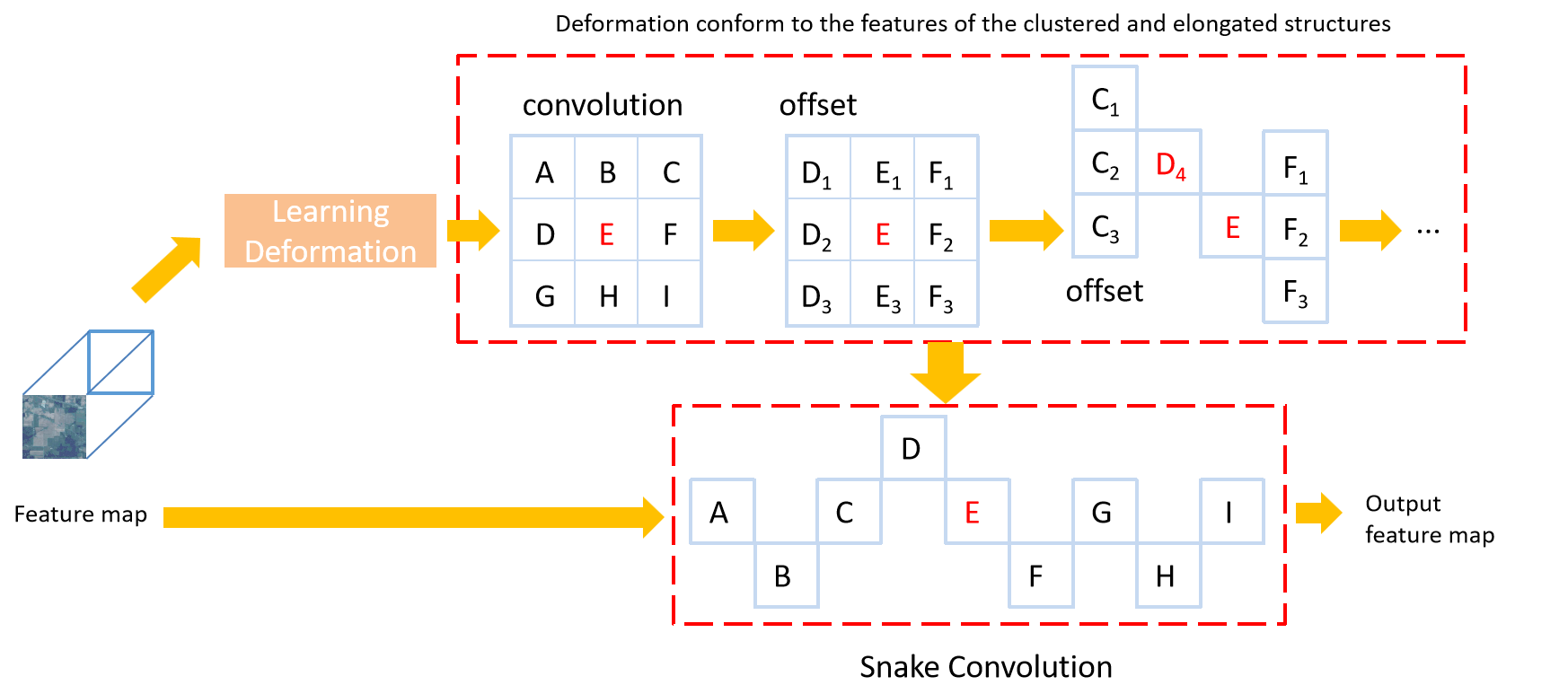}
\caption{DSCConv architecture diagram}
\label{fig:dsc_arch}
\end{figure}

\subsubsection{Multi-View Fusion}
We discusses the implementation of a multi-view feature fusion strategy to guide the model in complementing attention to key features from multiple perspectives. For each $K$, two feature maps $f_l(K_x)$ and $f_l(K_y)$ are extracted from the $x$-axis and $y$-axis at the $l$-th layer, expressed as:

\begin{equation}
f_l(K) = \Bigg\{
\underbrace{\sum_i w(K_i) \cdot f_l(K_i)}_{f_l(K_x)},
\underbrace{\sum_j w(K_j) \cdot f_l(K_j)}_{f_l(K_y)}
\Bigg\}
\end{equation}

where $w(K_i)$ denotes the weight at position $K_i$, and the features extracted by the convolutional kernel $K$ at the $l$-th layer are computed using an accumulation method.

Based on Equation (6), we extract $m$ sets of features as $\mathcal{T}_l$, which include different configurations of DSConv:

\begin{equation}
\mathcal{T}_l = \Bigg(
\underbrace{f_l(K_x), f_l(K_y)}_{\mathcal{T}_l^1},
\underbrace{f_l(K_x), f_l(K_y)}_{\mathcal{T}_l^2},
\cdots
\underbrace{f_l(K_x), f_l(K_y)}_{\mathcal{T}_l^m}
\Bigg)
\end{equation}

The fusion of multiple templates will inevitably introduce redundant noise. Therefore, during the training phase, we introduce a random dropout strategy $r_l$ to improve model performance and prevent overfitting without additional computational overhead. Equation (7) then becomes:

\begin{equation}
\begin{cases}
r_l \sim \text{Bernoulli}(p) \\
\hat{\mathcal{T}}_l = r_l \cdot \mathcal{T}_l \\
f_{l+1}(K) = \sum^{\lfloor m \times p \rfloor} \hat{\mathcal{T}}_l^t
\end{cases}
\end{equation}

where $p$ is the dropout probability, and $r_l$ follows a Bernoulli distribution. The optimal dropout strategy is preserved during training and guides the model to fuse key features during testing.

In our hyperspectral network architecture, the 3D DSConv combined with the multi-view fusion strategy ensures that while sampling spatial-spectral features, we can adaptively and dynamically optimize the network's feature representation capability. This achieves dynamic filtering in both spatial and spectral dimensions of hyperspectral data - focusing more on responses to key spatial structures while implementing identification and skipping of redundant spectral information.

\begin{figure}[h]
\centering
\includegraphics[width=0.9\linewidth]{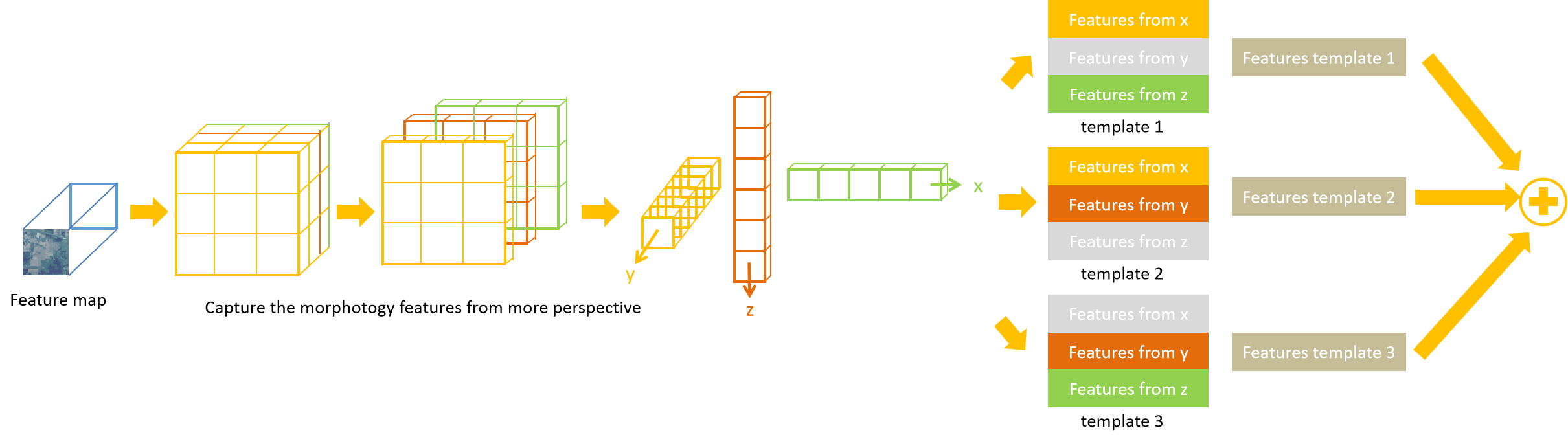}
\caption{Multi-view fusion strategy diagram}
\label{fig:fusion_strategy}
\end{figure}

\subsection{3D-CNN Framework for HSI Feature Extraction}

We modify original 3D-DenseNet in two aspects to simplify architecture and improve efficiency.

\begin{figure}[h]
\centering
\includegraphics[width=0.9\linewidth]{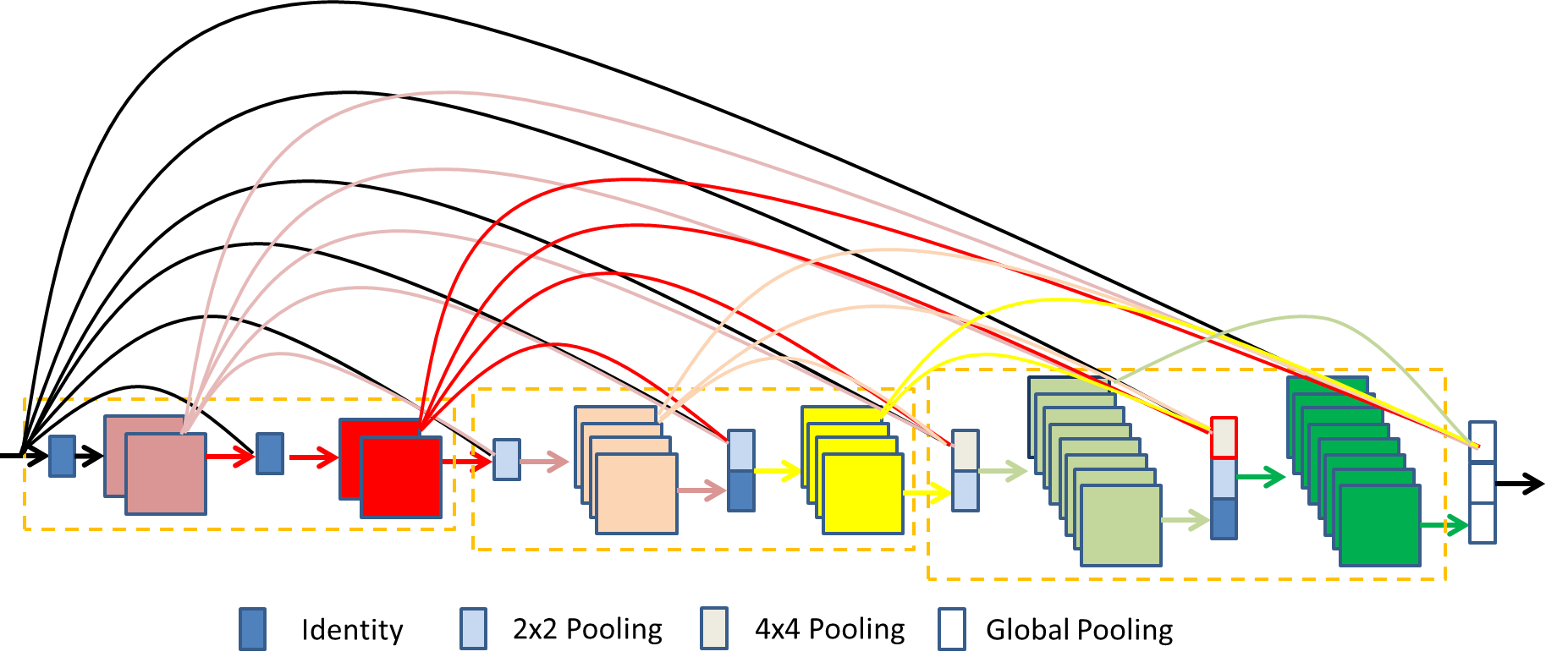}
\caption{The proposed DenseNet variant. It differs from the original DenseNet in two aspects: (1) layers with different resolution feature maps are also directly connected; (2) the growth rate doubles whenever the feature map size is downsampled (features generated in the third yellow dense block are significantly more numerous than those in the first block).}
\label{fig:densenet_variant}
\end{figure}

\subsubsection{Exponentially Increasing Growth Rate}
The original DenseNet architecture appends a constant number ($k$) of feature maps at each layer, where $k$ is termed the \textit{growth rate}. As demonstrated in \cite{huang2017densely}, deeper layers in DenseNet exhibit stronger dependencies on high-level features rather than low-level features. This observation motivates our architectural enhancement through strengthened short-range connections, achieved by progressively increasing the growth rate with network depth.Mathematically, we set the growth rate as $k=2^{m-1}k_0$, where $m$ is the dense block index and $k_0$ is a constant. This growth rate configuration introduces no additional hyperparameters. The ``increasing growth rate'' strategy allocates a larger proportion of parameters to the deeper layers of the model. While this approach significantly improves computational efficiency, it may potentially reduce parameter efficiency in certain scenarios. Depending on specific hardware constraints, trading one aspect for the other could be advantageous \cite{liu2017learning}..

\begin{figure}[h]
\centering
\includegraphics[width=0.9\linewidth]{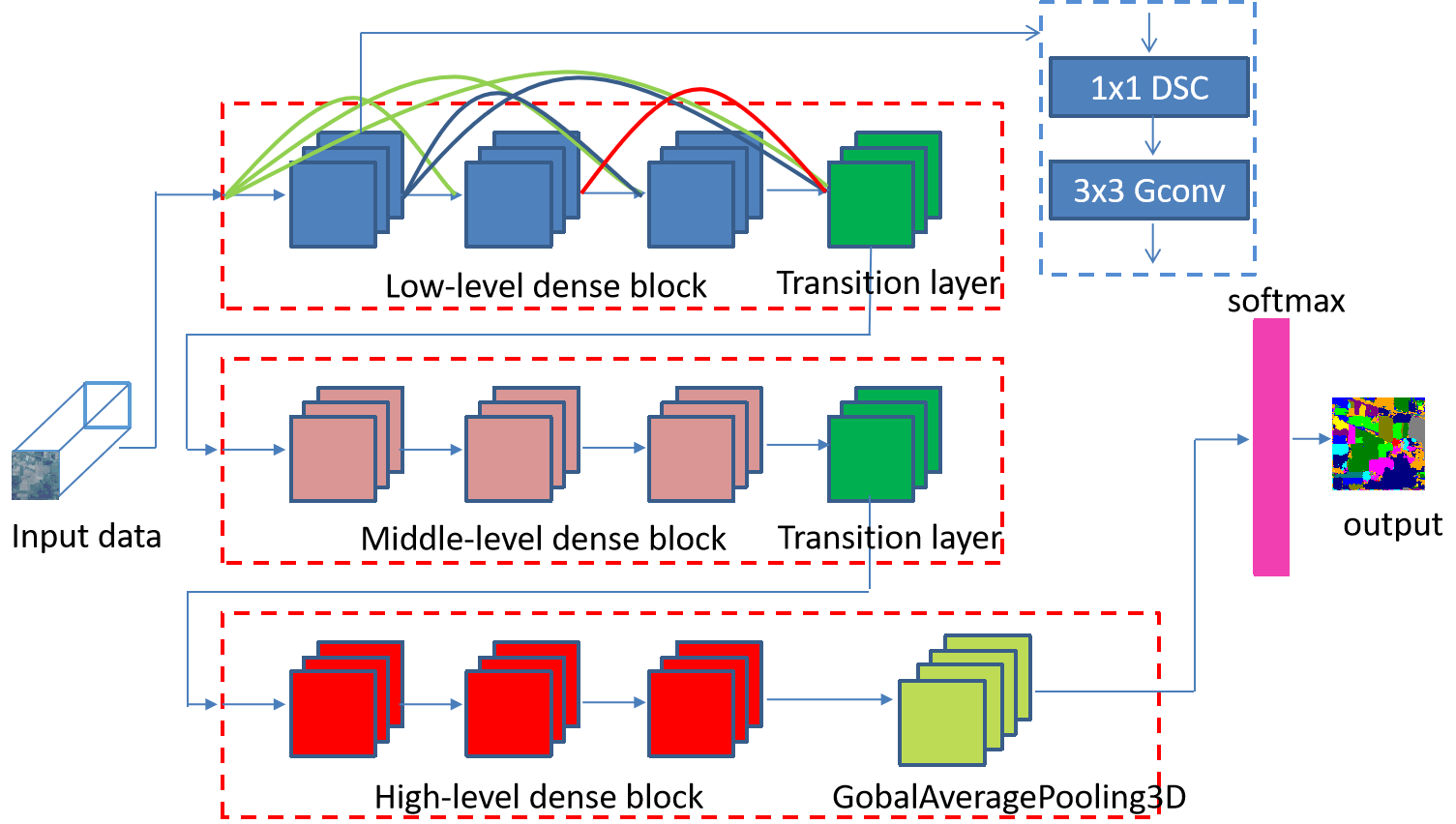}
\caption{SG-DSCNet architecture incorporating modified 3D-DenseNet framework}
\label{fig:sgdscnet_arch}
\end{figure}

\subsubsection{Fully Dense Connectivity}
To promote greater feature reuse compared to the original DenseNet architecture, we connect the input layer to all subsequent layers in the network, even those residing in different dense blocks (Fig.\ref{fig:densenet_variant}). Since dense blocks exhibit varying feature resolutions, we downsample feature maps with higher resolutions using average pooling when they are utilized as inputs to layers with lower resolutions.

\section{Experiments and Analysis}
To evaluate the performance of SG-DSCNet, we conducted experiments on three representative hyperspectral datasets: Indian Pines test site, Pavia University urban area, and Kennedy Space Center (KSC). Evaluation metrics include Overall Accuracy (OA), Average Accuracy (AA), and Kappa coefficient.

\subsection{Datasets}
\subsubsection{Indian Pines Dataset}
Collected in June 1992 by AVIRIS sensor over Northwestern Indiana, this dataset contains $145 \times 145$ pixels with 20\,m spatial resolution. Originally having 220 spectral bands ($0.4$--$2.5\,\mu\text{m}$), we removed 20 water absorption bands, retaining 200 bands for experiments. It contains 16 land cover classes.

\begin{figure}[t]
\centering
\includegraphics[width=0.6\textwidth]{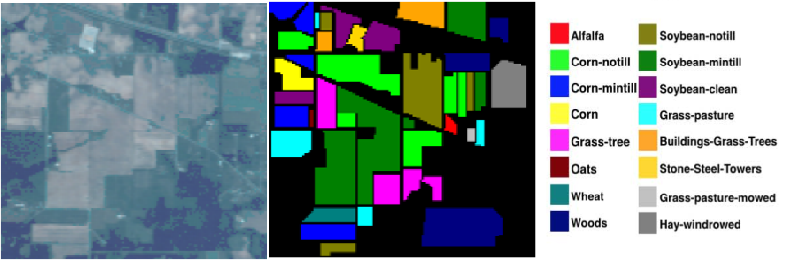}
\caption{False color image and ground-truth labels of Indian Pines}
\label{fig:indian_pines}
\end{figure}

\subsubsection{Pavia University Dataset}
Acquired in 2001 by ROSIS sensor over Pavia, Italy, this dataset has $610 \times 340$ pixels with 1.3\,m resolution. From 115 original bands ($0.43$--$0.86\,\mu\text{m}$), we removed 12 noisy bands, keeping 103 bands. It contains 9 urban land cover classes.

\begin{figure}[t]
\centering
\includegraphics[width=0.6\textwidth]{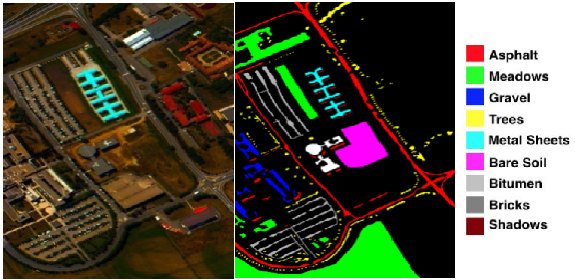}
\caption{False color image and ground-truth labels of Pavia University dataset}
\label{fig:pavia}
\end{figure}

\subsubsection{KSC Dataset}
Collected by AVIRIS in March 1996 over Kennedy Space Center, Florida, this dataset has 224 bands (10\,nm width, $400$--$2500\,\text{nm}$) with 18\,m resolution. After removing water absorption bands, 176 bands were used, containing 13 land cover classes.

\subsection{Experimental Analysis}
SG-DSCNet was trained for 80 epochs on all datasets using Adam optimizer on four 80GB A100 GPUs. We use a 3-stage base architecture (4,6,8 dense blocks) with growth rates 8,16,32, 4 heads, 3×3 group convolution (4 groups), gate\_factor=0.25, and compression=16.

\subsubsection{Data Partitioning Ratio}
For hyperspectral data with small sample sizes, the proportion of the training set significantly affects the stability and generalization ability of model performance. To thoroughly evaluate the sensitivity of data partitioning strategies, this study conducted a comparative analysis of model performance under different ratios of training, validation, and test sets. Experimental results show that when training samples are limited, adopting a 6:1:3 split ratio effectively balances the model's learning capacity and evaluation reliability—this configuration allocates 60\% of the samples for sufficient training, utilizes a 10\% validation set to implement an early stopping mechanism to prevent overfitting, and reserves 30\% for the test set to ensure statistically significant evaluation. SG-DSCNet adopted this 6:1:3 split ratio across the Indian Pines, Pavia University, and KSC benchmark datasets, with an 11×11 neighboring pixel block size to balance local feature extraction and the integrity of spatial contextual information.

\begin{table}[h]
\begin{minipage}{\textwidth}
\centering
\makeatletter
\def\@makecaption#1#2{%
    \vskip\abovecaptionskip
    \centering 
    \small #1: #2\par
    \vskip\belowcaptionskip
}
\makeatother
\caption{OA, AA and Kappa metrics for different training set ratios on the Indian Pines dataset}
\begin{adjustbox}{width=0.5\columnwidth}
\begin{tabular}{cccc}
\toprule
Training Ratio & OA & AA & Kappa \\
\midrule
2:1:7 & 92.33 & 74.17 & 91.24 \\
3:1:6 & 97.88 & 95.90 & 97.58 \\
4:1:5 & 97.93 & 98.08 & 97.65 \\
5:1:4 & 99.12 & 98.41 & 99.00 \\
6:1:3 & 99.77 & 99.78 & 99.74 \\
\bottomrule
\end{tabular}
\label{tab:indian_ratios}
\end{adjustbox}
    \end{minipage}

 \vspace*{10pt} 
 
\begin{minipage}{\textwidth}
\centering
\makeatletter
\def\@makecaption#1#2{%
    \vskip\abovecaptionskip
    \centering 
    \small #1: #2\par
    \vskip\belowcaptionskip
}
\makeatother
\caption{OA, AA and Kappa metrics for different training set ratios on the Pavia University dataset}
\begin{tabular}{cccc}
\toprule
Training Ratio & OA & AA & Kappa \\
\midrule
2:1:7 & 99.41 & 99.10 & 99.21 \\
3:1:6 & 99.53 & 99.39 & 99.39 \\
4:1:5 & 99.61 & 99.36 & 99.50 \\
5:1:4 & 99.92 & 99.89 & 99.90 \\
6:1:3 & 99.99 & 99.99 & 99.99 \\
\bottomrule
\end{tabular}
\label{tab:pavia_ratios}
        \end{minipage}

 \vspace*{10pt} 
 
\begin{minipage}{\textwidth}        
\centering
\makeatletter
\def\@makecaption#1#2{%
    \vskip\abovecaptionskip
    \centering 
    \small #1: #2\par
    \vskip\belowcaptionskip
}
\makeatother
\caption{OA, AA and Kappa metrics for different training set ratios on the KSC dataset}
\begin{tabular}{cccc}
\toprule
Training Ratio & OA & AA & Kappa \\
\midrule
2:1:7 & 96.84 & 95.75 & 96.48 \\
3:1:6 & 99.52 & 99.18 & 99.47 \\
4:1:5 & 98.69 & 98.26 & 98.54 \\
5:1:4 & 99.33 & 99.05 & 99.25 \\
6:1:3 & 99.67 & 99.50 & 99.64 \\
\bottomrule
\end{tabular}
\label{tab:ksc_ratios}
        \end{minipage}
\end{table}

\subsubsection{Neighboring Pixel Blocks}
The network performs edge padding on the input image of size 145×145×103, transforming it into a 155×155×103 image (taking Indian Pines as an example). On this 155×155×103 image, an adjacent pixel block of size M×N×L is sequentially selected, where M×N represents the spatial sampling size, and L denotes the full-dimensional spectrum. The original image is too large, which is not conducive to sufficient feature extraction through convolution, slows down the computation speed, increases short-term memory usage, and places high demands on the hardware platform. Therefore, processing with adjacent pixel blocks is adopted, where the size of the adjacent pixel block is an important hyperparameter. However, the range of the adjacent pixel block cannot be too small, as it may lead to an insufficient receptive field for the convolution kernel's feature extraction, resulting in poor local performance. As shown in Tables \ref{tab:indian_blocks}-\ref{tab:ksc_blocks}, on the Indian Pines dataset, as the pixel block size increases from 7 to 17, there is a noticeable improvement in accuracy. This trend is also evident in the Pavia University dataset. However, as the pixel block range continues to expand, the overall accuracy growth diminishes, exhibiting a clear threshold effect, which is similarly observed in the KSC dataset. Therefore, we selected a neighboring pixel block size of 15 for the Indian Pines dataset, 11 for the Pavia University dataset, and 17 for the KSC dataset.

\begin{table}[h]
\begin{minipage}{\textwidth}
\centering
\makeatletter
\def\@makecaption#1#2{%
    \vskip\abovecaptionskip
    \centering 
    \small #1: #2\par
    \vskip\belowcaptionskip
}
\makeatother
\caption{OA, AA and Kappa metrics for different block sizes on Indian Pines}
\begin{tabular}{cccc}
\toprule
Block Size (M=N) & OA & AA & Kappa \\
\midrule
7 & 98.11 & 96.48 & 97.85 \\
9 & 99.45 & 99.13 & 99.37 \\
11 & 99.77 & 99.78 & 99.74 \\
13 & 99.81 & 99.69 & 99.78 \\
15 & 99.90 & 99.75 & 99.89 \\
17 & 99.74 & 99.44 & 99.70 \\
\bottomrule
\end{tabular}
\label{tab:indian_blocks}
        \end{minipage}

 \vspace*{10pt} 
 
\begin{minipage}{\textwidth}
\centering
\makeatletter
\def\@makecaption#1#2{%
    \vskip\abovecaptionskip
    \centering 
    \small #1: #2\par
    \vskip\belowcaptionskip
}
\makeatother
\caption{OA, AA and Kappa metrics for different block sizes on Pavia University}
\begin{tabular}{cccc}
\toprule
Block Size (M=N) & OA & AA & Kappa \\
\midrule
7 & 99.53 & 99.41 & 99.38 \\
9 & 99.84 & 99.78 & 99.79 \\
11 & 99.99 & 99.99 & 99.99 \\
13 & 99.95 & 99.93 & 99.93 \\
15 & 99.94 & 99.84 & 99.92 \\
17 & 99.99 & 99.99 & 99.99 \\
\bottomrule
\end{tabular}
\label{tab:pavia_blocks}
        \end{minipage}

 \vspace*{10pt} 
 
\begin{minipage}{\textwidth}
\centering
\makeatletter
\def\@makecaption#1#2{%
    \vskip\abovecaptionskip
    \centering 
    \small #1: #2\par
    \vskip\belowcaptionskip
}
\makeatother
\caption{OA, AA and Kappa metrics for different block sizes on KSC dataset}
\begin{tabular}{cccc}
\toprule
Block Size (M=N) & OA & AA & Kappa \\
\midrule
7 & 98.72 & 97.56 & 98.57 \\
9 & 99.10 & 98.83 & 98.99 \\
11 & 99.68 & 99.50 & 99.64 \\
13 & 1 & 1 & 1 \\
15 & 99.90 & 99.75 & 99.89 \\
17 & 1 & 1 & 1 \\
\bottomrule
\end{tabular}
\label{tab:ksc_blocks}
        \end{minipage}
\end{table}

\subsubsection{Network Parameters}
We divided SG-DSCNet into two types: base and large. The Table \ref{tab:sgdscnet_config} below presents the OA, AA, and Kappa results tested on the Indian Pines dataset. In the large model, the number of dense blocks in each of the three stages is set to 14, resulting in a significantly larger parameter count compared to the base model. However, the accuracy does not exhibit a substantial proportional increase relative to our base version, showing a clear effect of diminishing marginal returns. For both the base and large versions used in the comparison, the training set ratio is 6:1:3, and the adjacent pixel block size is 11.

\begin{table}[h]
\centering
\makeatletter
\def\@makecaption#1#2{%
    \vskip\abovecaptionskip
    \centering 
    \small #1: #2\par
    \vskip\belowcaptionskip
}
\makeatother
\caption{SG-DSCNet model configurations and performance metrics}
\label{tab:sgdscnet_config}
\begin{tabular}{lccccc}  
\toprule
Model & Stages/Blocks & Growth Rate & OA & AA & Kappa \\
\midrule
SG-DSCNet-base & 4,6,8 & 8,16,32 & 99.90 & 99.75 & 99.89 \\
SG-DSCNet-large & 14,14,14 & 8,16,32 & 99.93 & 99.77 & 99.93 \\
\bottomrule
\end{tabular}
\end{table}

\subsection{Experimental Results and Analysis}
\begin{figure}[h]
\centering
\includegraphics[width=0.95\linewidth]{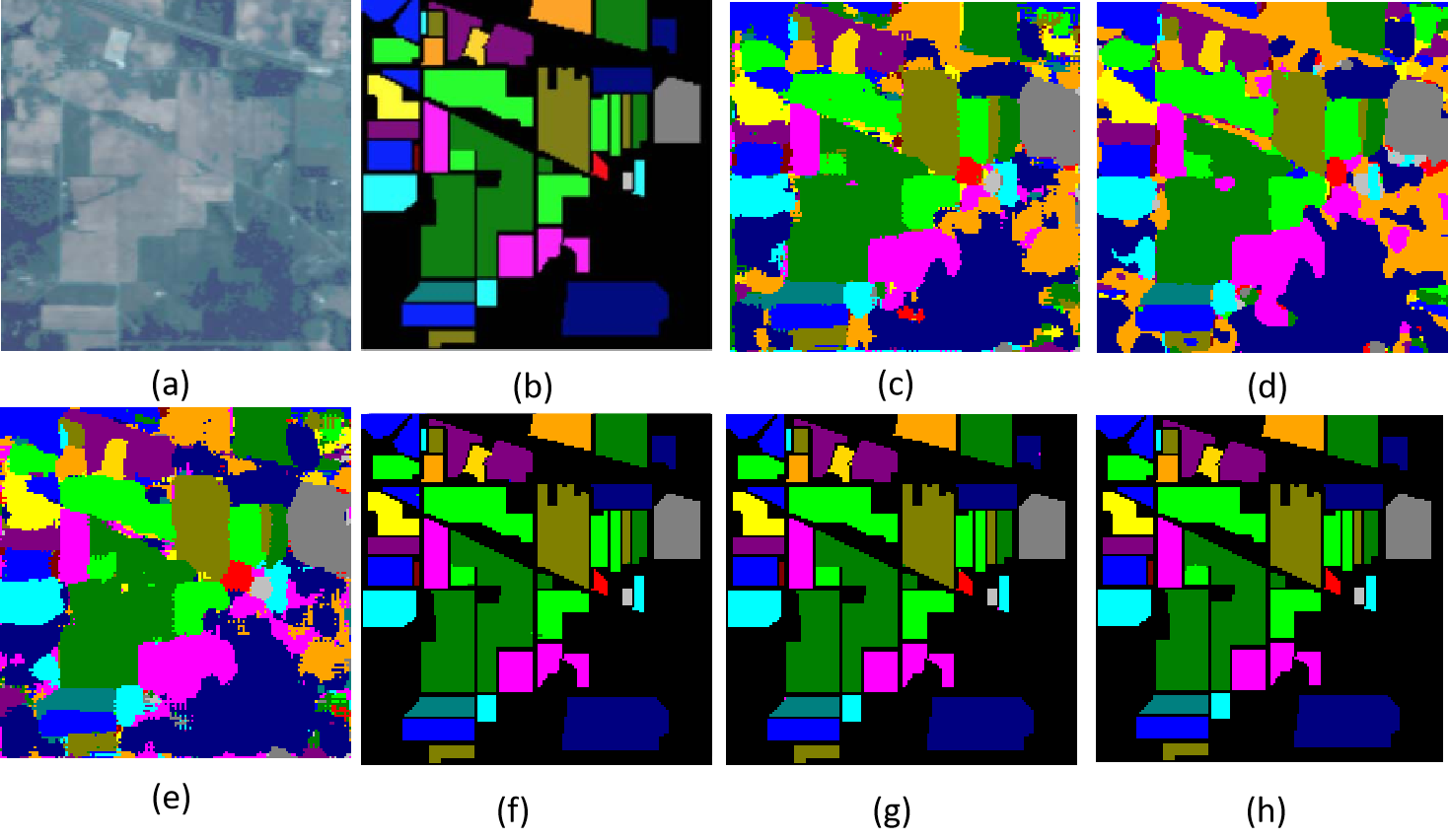}
\caption{Classification results on Indian Pines dataset: (a) Sample image, (b) Ground truth labels, (c)-(i) Results from SSRN, 3D-CNN, 3D-SE-DenseNet, LGCNet, DGCNet and SG-DSCNet}
\label{fig:indian_results}
\end{figure}

On the Indian Pines dataset, the input size for SG-DSCNet is $17\times17\times200$; on the Pavia University dataset, it is $17\times17\times103$; and on the KSC dataset, it is $17\times17\times176$. The experiment compared SSRN, 3D-CNN, 3D-SE-DenseNet, Spectralformer, LGCNet, DGCNet, and our SG-DSCNet-base. As shown in Tables \ref{tab:indian_results}-\ref{tab:pavia_results}, SG-DSCNet achieved leading accuracy overall on both datasets. Fig.\ref{fig:indian_results} illustrates the loss and accuracy changes on the training and validation sets, demonstrating that the loss function converges very quickly and the accuracy increases in a stable manner.

\begin{table*}[!ht]
\begin{minipage}{\textwidth}
\centering
\makeatletter
\def\@makecaption#1#2{%
    \vskip\abovecaptionskip
    \centering 
    \small #1: #2\par
    \vskip\belowcaptionskip
}
\makeatother
\caption{Classification accuracy comparison (\%) on Indian Pines dataset}
\label{tab:indian_results}
\resizebox{\textwidth}{!}{
\begin{tabular}{@{}lccccccc@{}}
\toprule
Class & SSRN & 3D-CNN & 3D-SE-DenseNet & Spectralformer & LGCNet & DGCNet & SG-DSCNet \\
\midrule
1 & 100 & 96.88 & 95.87 & 70.52 & 100 & 100 & 100 \\
2 & 99.85 & 98.02 & 98.82 & 81.89 & 99.92 & 99.47 & 99.76 \\
3 & 99.83 & 97.74 & 99.12 & 91.30 & 99.87 & 99.51 & 100 \\
4 & 100 & 96.89 & 94.83 & 95.53 & 100 & 97.65 & 100 \\
5 & 99.78 & 99.12 & 99.86 & 85.51 & 100 & 100 & 100 \\
6 & 99.81 & 99.41 & 99.33 & 99.32 & 99.56 & 99.88 & 100 \\
7 & 100 & 88.89 & 97.37 & 81.81 & 95.83 & 100 & 100 \\
8 & 100 & 100 & 100 & 75.48 & 100 & 100 & 100 \\
9 & 0 & 100 & 100 & 73.76 & 100 & 100 & 100 \\
10 & 100 & 100 & 99.48 & 98.77 & 99.78 & 98.85 & 100 \\
11 & 99.62 & 99.33 & 98.95 & 93.17 & 99.82 & 99.72 & 99.86 \\
12 & 99.17 & 97.67 & 95.75 & 78.48 & 100 & 99.56 & 100 \\
13 & 100 & 99.64 & 99.28 & 100 & 100 & 100 & 100 \\
14 & 98.87 & 99.65 & 99.55 & 79.49 & 100 & 99.87 & 100 \\
15 & 100 & 96.34 & 98.70 & 100 & 100 & 100 & 100 \\
16 & 98.51 & 97.92 & 96.51 & 100 & 97.73 & 98.30 & 96.43 \\
\midrule
OA & 99.62$\pm$0.00 & 98.23$\pm$0.12 & 98.84$\pm$0.18 & 81.76 & 99.85$\pm$0.04 & 99.58 & 99.90 \\
AA & 93.46$\pm$0.50 & 98.80$\pm$0.11 & 98.42$\pm$0.56 & 87.81 & 99.53$\pm$0.23 & 99.55 & 99.75 \\
K & 99.57$\pm$0.00 & 97.96$\pm$0.53 & 98.60$\pm$0.16 & 79.19 & 99.83$\pm$0.05 & 99.53 & 99.89 \\
\bottomrule
\end{tabular}
}
        \end{minipage}

 \vspace*{10pt} 

\begin{minipage}{\textwidth}

\centering
\makeatletter
\def\@makecaption#1#2{%
    \vskip\abovecaptionskip
    \centering 
    \small #1: #2\par
    \vskip\belowcaptionskip
}
\makeatother
\caption{Classification accuracy comparison (\%) on Pavia University dataset}
\label{tab:pavia_results}
\resizebox{\textwidth}{!}{
\begin{tabular}{@{}lcccccc@{}}
\toprule
Class & SSRN & 3D-CNN & 3D-SE-DenseNet & Spectralformer & LGCNet & SG-DSCNet \\
\midrule
1 & 89.93 & 99.96 & 99.32 & 82.73 & 100 & 99.95 \\
2 & 86.48 & 99.99 & 99.87 & 94.03 & 100 & 100 \\
3 & 99.95 & 99.64 & 96.76 & 73.66 & 99.88 & 100 \\
4 & 95.78 & 99.83 & 99.23 & 93.75 & 100 & 100 \\
5 & 97.69 & 99.81 & 99.64 & 99.28 & 100 & 100 \\
6 & 95.44 & 99.98 & 99.80 & 90.75 & 100 & 100 \\
7 & 84.40 & 97.97 & 99.47 & 87.56 & 100 & 100 \\
8 & 100 & 99.56 & 99.32 & 95.81 & 100 & 100 \\
9 & 87.24 & 100 & 100 & 94.21 & 100 & 100 \\
\midrule
OA & 92.99$\pm$0.39 & 99.79$\pm$0.01 & 99.48$\pm$0.02 & 91.07 & 99.99$\pm$0.00 & 99.99 \\
AA & 87.21$\pm$0.25 & 99.75$\pm$0.15 & 99.16$\pm$0.37 & 90.20 & 99.99$\pm$0.01 & 99.99 \\
K & 90.58$\pm$0.18 & 99.87$\pm$0.27 & 99.31$\pm$0.03 & 88.05 & 99.99$\pm$0.00 & 99.99 \\
\bottomrule
\end{tabular}
}
        \end{minipage}
\end{table*}

\begin{figure}[h]
\centering
\includegraphics[width=0.9\linewidth]{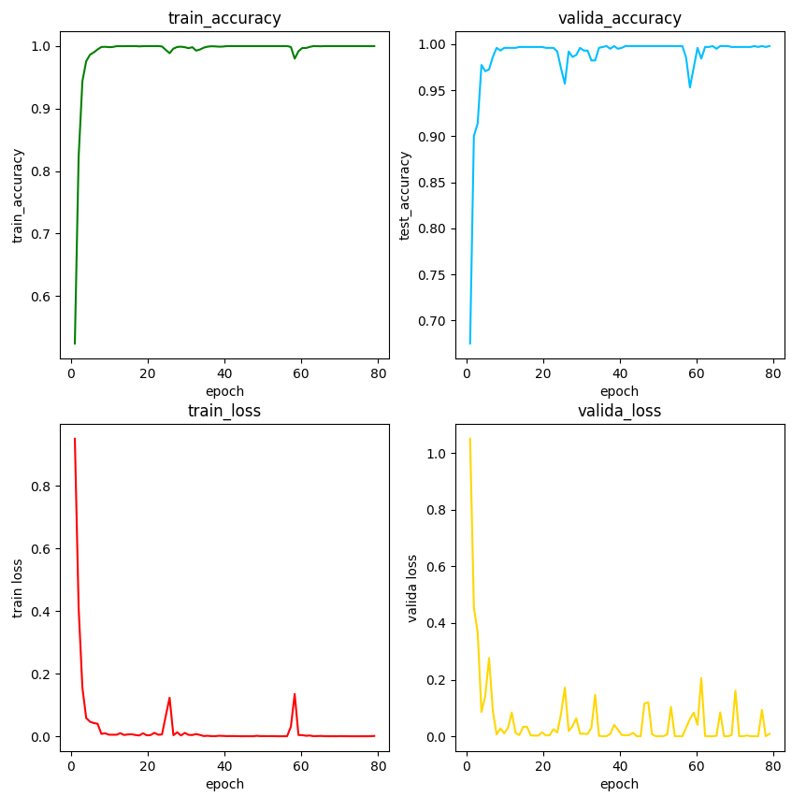}
\caption{Training curves of SG-DSCNet showing loss and accuracy evolution}
\label{fig:training_curves}
\end{figure}

\section{Conclusion}
This paper introduces DSC Conv on 3D convolution kernels. This module incorporates deformable offsets into standard 3D convolution, enhancing the flexibility of the convolution kernel and enabling it to adaptively focus on the geometric features of small clustered regions and elongated branching structures. By constraining the self-learning process, DSCConv avoids the disordered roaming issue of traditional deformable convolution in the perception area, thereby more accurately capturing key spatial structures in hyperspectral imagery. Additionally, a multi-view feature fusion strategy generates multiple morphological kernel templates to extract target features from different perspectives and perform efficient fusion, further improving the model’s ability to represent joint spatial-spectral information. This dynamic design allows SG-DSCNet to flexibly adapt to the feature demands of different spatial regions without relying on a single static convolution kernel, while effectively skipping redundant information and reducing computational complexity. SG-DSCNet further leverages the 3D-DenseNet architecture to extract spatial structures and spectral information of more critical features, providing an efficient solution for hyperspectral image classification. It successfully addresses the challenges posed by sparse object distribution and spectral redundancy.

{\small
\bibliographystyle{template}
\bibliography{template}
}

\end{document}